\documentclass[a4paper,fleqn]{cas-dc}

\usepackage[numbers]{natbib}
\usepackage[ruled,vlined]{algorithm2e}
\usepackage{graphicx}
\usepackage{amsmath}
\usepackage{hyperref}
\def\tsc#1{\csdef{#1}{\textsc{\lowercase{#1}}\xspace}}
\tsc{WGM}
\tsc{QE}
\tsc{EP}
\tsc{PMS}
\tsc{BEC}
\tsc{DE}

\begin{document}
\shorttitle{Dynamic Crowd Clustering}
\shortauthors{A.B.M Wijaya et~al.}

\title [mode = title]{Efficient Dense Crowd Trajectory Prediction Via Dynamic Clustering}                      

\tnotetext[1]{This research was funded by the Center of Higher Education Funding and Assessment (PPAPT), the Indonesian Ministry of Higher Education and Research, the Indonesian Education Scholarship (BPI), and the Indonesian Endowment Fund for Education (LPDP).}

\author[1]{Antonius Bima Murti Wijaya}
\ead{a.wijaya.1@research.gla.ac.uk}
\author[1]{Paul Henderson}
\ead{paul.henderson@glasgow.ac.id}
\author[1]{Marwa Mahmoud}
\ead{marwa.mahmoud@glasgow.ac.id}

\affiliation[1]{organization=School of Computing Science, University of Glasgow,
                addressline={University Avenue}, 
                city={Glasgow},
                postcode={G12 8QQ}, 
                state={Scotland},
                country={United Kingdom}}

\begin{abstract}
Crowd trajectory prediction plays a crucial role in public safety and management, where it can help prevent disasters such as stampedes. Recent trajectory prediction works address the problem by predicting individual trajectories and considering surrounding objects based on manually annotated data. However, these approaches tend to overlook dense crowd scenarios, where the challenges of automation become more pronounced due to the massiveness, noisiness, and inaccuracy of the tracking outputs, resulting in high computational costs. To address these challenges, we propose and extensively evaluate a novel cluster-based approach that groups individuals based on similar attributes over time, enabling faster execution through accurate group summarisation. Our plug-and-play method can be combined with existing trajectory predictors by using our output centroid in place of their pedestrian input. We evaluate our proposed method on several challenging dense crowd scenes. We demonstrated that our approach leads to faster processing and lower memory usage when compared with state-of-the-art methods, maintaining accuracy with minimal performance degradation. \url{https://github.com/bimamurti/CrowdCluster}
\end{abstract}
\maketitle
\section{Introduction}

\label{sec:intro}
Understanding crowd behaviour effectively requires methods capable of rapid analysis, particularly in applications involving trajectory prediction within dense, crowded environments to prevent accidents such as stampedes or overcrowding. In crowds, individuals often adjust their movements to align with those around them, naturally following the behaviour of their neighbours to avoid collisions. 

Prior work on trajectory prediction has primarily addressed scenarios with lower pedestrian density and has relied on manually annotated tracking data that provides reliable input \cite{Salzmann2020, Xugroupnet2022, Alahi2016}. This dependency becomes a challenge when the scenario moves to a dense crowd, where the pedestrian's head is the most visible body part among many individuals, which reduces the object information. Consequently, the tracking accuracy is typically reduced, which affects the trajectory prediction accuracy, and its computational cost increases \cite{Sundararaman2021,Dehghan2018}, particularly for methods that use neighbourhood or group data. Recently, the Graph-based Conditional Variational Recurrent Neural Network (GCVRNN) \cite{Xu2023} was introduced to handle this missing trajectory problem with imputation using generative methods. However, imputation in the context of crowd tracking noise is more challenging, since the issue is not only missing tracks but also identity switches or changes.

To address these challenges, this paper proposes a clustering strategy inspired by how humans mimic other's trajectory behaviour to move through spaces efficiently \cite{Dehghan2018,Zhou2022}. This clustering approach reduces the execution time while maintaining comparable accuracy by reducing sensitivity to disappearance caused by re-identification failures. The idea is to group several pedestrians into a cluster during tracking, then use the cluster itself as a substitute for those pedestrians. Instead of operating on individual pedestrians, the trajectory predictor now operates on cluster centroids. This clustering technique also reduces privacy leakage from tracking, as it uses aggregated cluster data instead of individual tracking data for input into trajectory predictions.

We evaluate clustering performance using membership accuracy to assess how well the proposed clusters represent pedestrians in a dynamic crowd environment. We then combine our method with several different trajectory predictors and show that our approach maintains prediction accuracy while significantly reducing computational cost.

\section{Related Work}
\subsection{Pedestrian Tracking}
Pedestrian tracking is typically based on object detection methods and aims to track individuals accurately over time as they move through the scene \cite{Rudenko2020}. This is mostly achieved by representing the crowd with a statistical individual view, network, or graph, which is extracted from images with object detection methods \cite{Corbetta2023,Warren2024}. Linear regression and quadratic programming \cite{Dehghan2018} showed promising performance on pedestrian tracking. However, it does not inherit generalisation ability from the object detection base, which is less practical. SORT (Simple Online and Realtime Tracking)\cite{Bewley2016} became a popular tracking algorithm based on detection. Advanced work tried to employ deep learning on SORT to understand the appearance features by employing deep appearance descriptors for re-identification when the object is lost in the sequence \cite{wojke2017,Aharon2022,Zhang2022, Cao2023}. HeadHunter-T was introduced as a response to object detection for head tracking \cite{Sundararaman2021} and could successfully tackle many of the head tracking challenges. It incorporated appearance descriptors using the colour histogram to improve the re-identification performance. However, models that depend mainly on object detection results still inherently suffer from identity switches (ID-switch) and tracking loss, especially on low-resolution visual data.
\subsection{Pedestrian Trajectory Prediction}
Trajectory prediction involves forecasting the future positions of an object or agent based on its current trajectory, motion patterns, and environmental context. Alahi \emph{et al.} \cite{Alahi2016} introduced Social LSTM (Long Short-Term Memory) as the first approach that considered social data to improve trajectory prediction. When tested in a small crowd scenario, the Social LSTM concept proved capable of being adapted to predict various pedestrian movements even in a multiple-camera scenario \cite{Styles2020} \cite{Styles2022}. To improve the accuracy, Social GAN, which is based on a generative model, was introduced \cite{Gupta2018}. Other works also combined SocialLSTM with optical flow to predict pedestrian movements \cite{Wang2021,Chen2020,Styles2020} and achieved better accuracy. However, training this network could be very excessive for dense crowd scenarios since, for every pedestrian, it needs to consider all of their neighbour's data grids.

Trajectron then introduced an advanced network structure \cite{Salzmann2020} by combining a deep generative model of CVAEs (Conditional Variational Autoencoders), LSTM, and a Spatiotemporal Graphical Structure. This algorithm showed improvement in both ADE and FDE results compared to previous works. Trajectron++ extended this Trajectron framework to handle multimodality, facilitating its application in robotics navigation \cite{Salzmann2020}. Built upon the Trajectron++ framework, Grouptron \cite{Zhou2022} achieved significant performance improvements on ADE and FDE that focus on the pedestrian dataset. Grouptron incorporated group information collected from pedestrian clusters to enhance performance. It employed levelling group data extraction based on the cluster group. The incorporation of group information in Grouptron brought additional data for the model, which led to a heavier computational load.

With the same cluster utilisation, SocialVAE\cite{XuVAE2022} exploited the use of RNN and latent variables with a clustering algorithm to optimise the output. Furthermore, this method employed clustering as a post-processing method to improve its prediction accuracy. Even though it showed better performance than its predecessors, SocialVAE still suffered from memory consumption since it needs to store all of the neighbourhood data at the initialisation phase, and reduced trajectory prediction speed since it has to cluster the output. Groupnet\cite{Xugroupnet2022} came as a plug-and-play pedestrian grouping to extract group behaviour with a Multi-scale graph based on their neural interaction.  Groupnet also developed further with a transformer-based model into MART(Multiscale Relational Transformer) \cite{Lee2024}, which showed a significant performance improvement. Trajectron++, SocialVAE, and MART raise awareness of the other agents surrounding the model, which made the model learn bigger data. 
\subsection{Dynamic Clustering for Pedestrians}
Dynamic clustering is an advanced clustering technique that was introduced as a general concept for dealing with evolving environments. There are three main activities in dynamic clustering, namely: adding, removing, and updating \cite{Gu22}. DyClee is considered the latest development in this dynamic clustering \cite{roa2019}. It employs two stages of clustering: distance-based clustering followed by density-based clustering. Initially, DyClee clusters the data using either k-means or hierarchical clustering to initialise the first stage, and it describes the level of density inside a cluster as well. If new data comes, it compares nearby clusters; if it's reachable, then it finds the minimum distance. The density-based clustering evaluates the medium- to high-density clusters and connects the overlapping clusters. However, it has no mechanism to detect when a member leaves the current cluster; in a pedestrian-crowded case, people could leave a cluster if they change their trajectory or velocity. Therefore, an additional step is necessary to address this dynamic behaviour.

\section{Method}

\begin{figure*}
  \centering
  \includegraphics[width=\textwidth,keepaspectratio]{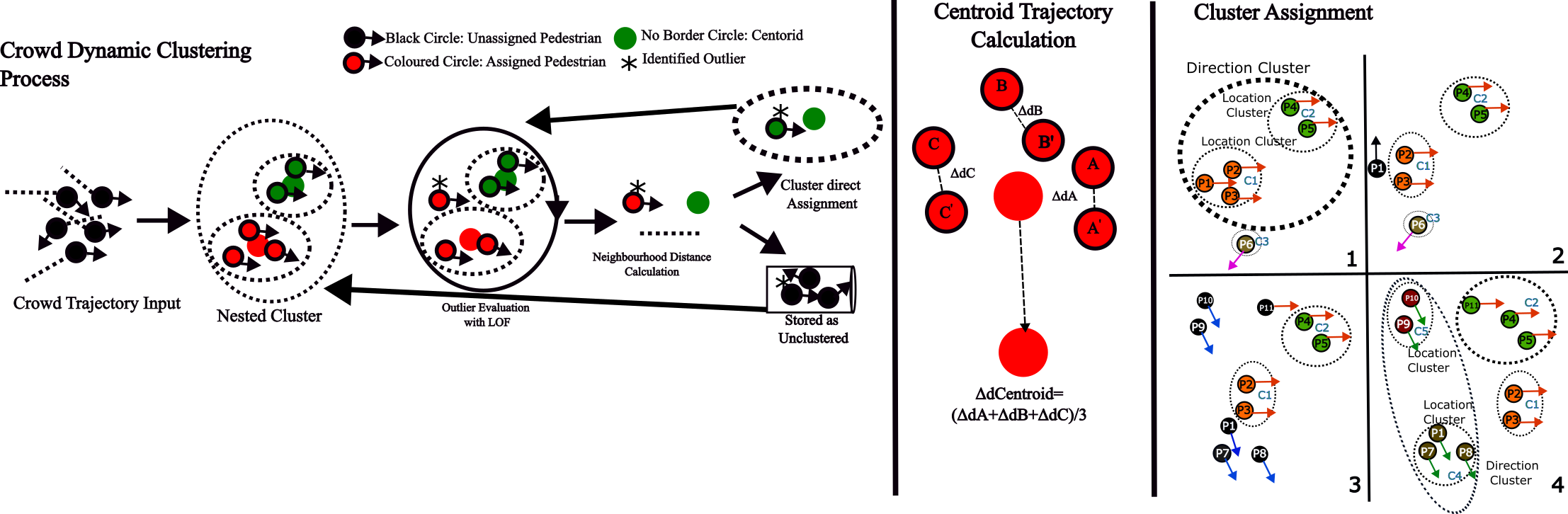}
    \vspace{-16pt}
   \caption{Our proposed dynamic clustering for dense crowds. The process starts with initialisation with a nested agglomerative cluster, evaluating the direction every 10 frames with LOF (Local Outlier Factor), and calculating the centroids. If LOF identifies outliers, then each outlier will be assigned to another nearby cluster or stored on an unassigned list. The process will execute the nested cluster again when a member of the unassigned list reaches a certain value. The Centroid Trajectory Calculation section shows how the cluster calculates its centroid trajectory accelerations based on the average deviation value from its membership value. The Cluster Assignment section illustrates the dynamic processing of clustering by the algorithm.  }
  \label{fig:map}
\end{figure*}

Our dynamic clustering method inputs a set of time-varying pedestrian locations, typically from an existing tracker. The clustering process produces a centroid location for each cluster in every frame. These centroid locations then replace individual pedestrian positions as input to an existing trajectory predictor, which learns to generate future trajectories.
This approach reduces training and inference time and memory usage for the prediction model, and is robust to noisy tracking input.
In Sec.~\ref{sec:clustering-strategy} we describe our dynamic clustering method for grouping pedestrians. In Sec.~\ref{sec:centroid-calculation}, we provide details on how cluster centroid tracks are defined even when pedestrians are added and removed dynamically.

\subsection{Clustering Strategy}
\label{sec:clustering-strategy}
 Our approach consists of two stages: nested distance-based clustering \cite{roa2019} for direction and distance, followed by a grouping stage that continuously re-evaluates cluster membership. The centroid of each cluster then represents its members' trajectories, and substitutes them as inputs for trajectory prediction.
Our dynamic clustering is based on two main input features: the direction angle and the location. $P$ represents the information for one pedestrian, containing: location ($x,y$), direction angle ($\theta$), direction vector ($vx,vy$), and cluster id ($Cid$):
\begin{equation}
\label{eq:input}
P_{nt}=
\begin{Bmatrix}
  x_{nt} ;
  y_{nt} ;
  \theta_{nt};
  vy_{nt} ;
  vx_{nt} ;
  Cid  
\end{Bmatrix}
\end{equation}
where $n$ indexes different individuals and $t$ is the time or frame number.

The direction vector is calculated by subtracting the previous and current location vectors; the direction angle is the arctangent of its components. The distance direction is calculated with the smallest angular distance. The direction angle ($\theta$) is extracted from the current and previous location. Other works \cite{Styles2020, Alahi2016, Ivanovic2019} used a directional vector as their feature for trajectory prediction. However, our clustering utilises the direction angle because we further used the smallest angular distance to calculate distance in direction.  

Given the trajectory data after several frames (see Figure \ref{fig:map} left), our method begins with clustering all of the detected pedestrians with direction-based clustering \cite{roa2019}, which is based on agglomerative clustering \cite{roa2019,Castellano2022}.  For each formed cluster, we apply another agglomerative clustering to its members based on the location of the pedestrian for each formed direction cluster as a nested cluster. The direction feature threshold ensures that the cluster method does not assign two deviating angles to the same cluster; therefore, it should not be more than 180. The location feature constraint limits the membership of the cluster under the location distance threshold. The threshold determines how far people can be included in one cluster and also how far the trajectories are allowed to deviate. It means the higher the value of the threshold, the more groups are created, which means less output data and a faster process. However, the drawback is that it has more deviations that affect the accuracy. In our experiment, we use 120px for distance and 60 for direction as a fair threshold; for a denser crowd, we can reduce the threshold.

An example is illustrated in Figure \ref{fig:map}. In the cluster assignment part, section 1, all pedestrians are clustered based on their direction. P1 to P5 are clustered into 1 direction cluster, and based on this cluster, it was re-clustered again into two location-based clusters, which were defined as cluster C1 and C2. Meanwhile, P6 is directly defined as C3 because it has no other members.

This initial process groups pedestrians coherently for a small number of frames, but it is insufficient for longer scenarios where the pedestrian location, direction, and numbers change over time \cite{Yang2018}. To facilitate this, we employ cluster evaluation through LOF \cite{Breunig2000}  every 10 frames to evaluate the membership based on its features. This 10-frame evaluation process results in a gap in the pedestrian count. The number of pedestrian members in the cluster is always lower than the ground truth dataset, but it eventually synchronises its position as long as all data is assigned to existing clusters or new clusters after the evaluation process. In the evaluation process, the LOF function evaluates the membership score of a cluster based on direction and location. In Figure~\ref{fig:map} `cluster assignment ' step, P1 is removed from C1 after the LOF algorithm identifies it as an anomaly. As no nearby cluster is available, P1 remains unassigned and is placed in a temporary list.

The algorithm then assigns the detected anomalies to the nearest cluster based on the direction and their nearby location, based on a specific location threshold. To calculate the similarity of the direction, the algorithm applies the Smallest Angular Distance. If there is no nearby cluster, the data is stored on a temporary list for future clustering. If there are already 5 or more pedestrians on the list, then it is clustered in the same way as the initial cluster using the nested cluster approach. New data is also assigned based on neighbouring clusters and put into the temporary list if they do not find any clusters nearby. These steps represent the adding and removing process of dynamic clustering \cite{Gu22}. Sections 3 and 4 in Figure \ref{fig:map}, pedestrian P11, which exhibits high similarity with C2, is assigned to that cluster. Meanwhile, P1, P7, P8, P9, and P10 undergo a nested clustering process similar to that in Section 1, forming clusters C4 and C5.

\subsection{Centroid Calculation}
\label{sec:centroid-calculation}
The crucial process for forming cluster trajectories is to determine the centroid feature values. Our approach requires a similar feature vector of the centroid of the pedestrian; therefore, we determine it as follows: 

\[
C_{it}=
\begin{bmatrix}
  X_{it} ; 
  Y_{it} ;
  \theta_{it}
\end{bmatrix}
\]
where $C$ represents the centroids, $i$ represents the cluster number, and $t$ represents the frame number. For the vector feature, $X$ and $Y$ are cluster locations, $\theta$ is the cluster direction angle calculated using direction vectors $vx$ and $vy$. The initial cluster (the first frame) locations are determined by calculating the average value of the cluster member's locations (see equation \ref{eq:clusteraverage}).
\begin{equation}
\label{eq:clusteraverage}
C_{0}  = \frac{\sum_{j=0}^{n} C_{jt}}{n}, \quad 
\end{equation}
where $n$ is the number of pedestrians in every cluster.

Using the average value for the rest of the frame would have made the cluster trajectory not smooth, since it caused it to have abrupt displacements that may occur due to the dynamic membership of the cluster and noise from the tracking methods.  Drawing inspiration from neural networks, which utilise the delta weight function to iteratively adjust weights, we adopt a similar approach. Every 10 frames, the centroid's location is determined by averaging the differences between the current locations of pedestrian members and their positions in the previous frame. Equation \ref{eq:clusterloc} shows the centroid location calculation ($Cp_{it}$), and centroid direction calculation $V{it}$:
\begin{equation}
\Delta{Cp_{it}}=\frac{\sum_{j=0}^{n}(Cp_{jt} - Cp_{jt-1})}{n}     
\end{equation}
\begin{equation}
\label{eq:clusterloc}
Cp_{it}=Cp_{it-1}+\Delta{Cp_{it}}; \quad
V{it}=Cp_{it-1}-Cp_{it}
\end{equation}
where
$Cp_{it}=[X_{it} ; 
  Y_{it}]$. We recalculate the direction vector and angles from the newest value of the cluster's location rather than using the previous direction to reduce cumulative error.

\section{Experiments}

We conducted two sets of experiments. First, we evaluated the clustering performance as a representation of pedestrian behaviour (Table~\ref{tab:resev}), aiming to compare the clustering performance between high-quality tracking data from the ground truth and automated tracking data generated by HeadHunter-T. Second, we evaluated the performance of the clustering approach in assisting trajectory predictors, specifically Trajectron++, SocialVAE, and MART (Tables~\ref{tab:trajectorypred}, \ref{tab:trajectorypredvae} \& \ref{tab:trajectorypredMART}).

\subsection{Evaluation Metrics}
\label{sec:eval-metrics}

Our aim is to ensure that each cluster is an accurate representation of its members, yet robust to members who lose tracking. Clusters should also exhibit a smooth path and effectively compress the pedestrian data. We therefore define several evaluation metrics to quantify and evaluate these aspects:

  \subsubsection{Cluster Trajectory Errors Occurrence (CTEO):} 
  
  Pedestrian trajectories should appear natural, meaning that no sudden displacements appear in their paths, which could disrupt the trajectory continuity. CTEO counts the percentage value of the noticeable path deviation occurrence of every cluster's trajectories. It counts the number of trajectory errors in each cluster and divides that by the total number of clusters, as shown in equation \ref{eq:CTEO}.
\[distl_t={\sqrt{(x_t-x_{t-1})^2+(y_t-y_{t-1})^2}}\]

\begin{equation}
\operatorname{distd}_t
=
\left|
\left(
\left(\theta_t - \theta_{t-1} + 180\right) \bmod 360
\right)
- 180
\right|
\end{equation}
 \begin{equation}
 \label{eq:CTEO}
 TEO=\frac{1}{f_i}\sum_{t=0}^{f} 1 [dist_t > T] ;\quad CTEO=\frac{1}{n}\sum_{i=0}^{n} TEO_i
 \end{equation}
$dist$ is the distance between the location($distl_t$) or direction ($ distd_t$)of a centroid between its frames, $T$ is a threshold for noticeable distance, $n$ is the cluster number and $f$ is the frame number on each centroid.
  \subsubsection{Cluster Trajectory Errors Length (CTEL)} The Length of the error shows how much the error affects the entire cluster's trajectory. CTEL counts the percentage value of the noticeable distance deviation length of every cluster's trajectory. It counts every cluster trajectory error length and divides it by the number of clusters, as shown in equation \ref{eq:CTEL}.
\begin{equation}
\begin{split}
 \label{eq:CTEL}
\text{TEL}= \sum_{t=1}^{f} \left[ \begin{array}{ll}
dist_t & \text{if } distl_t > T \\
0 & \text{otherwise}
\end{array} \right] ;
\\CTEL=\frac{1}{n}\sum_{i=0}^{n} TEL_{[i]}
\end{split}
\end{equation}
  \subsubsection{ Cluster Member Distance Deviations (CMDD)} This metric calculates the average value of the distance from every cluster member to its centroid, and calculates the average value for all clusters and frames. This metric only considers the centroids that have more than two members and shows how far the member is from its centroid, as shown in equation \ref{eq:CMDD}.
\begin{equation}
\begin{split}
 \label{eq:CMDD}
\beta=\left( \frac{1}{z_i} \sum_{j=0}^{z_i}\left| ped_t - C_t \right| \right);\quad
{\text{CMDD}} = \frac{1}{f,nc} \sum_{t=0}^{f}\sum_{i=0}^{nc} \beta_{t,i}
\end{split}
\end{equation}
where $z_i$ is the number of pedestrians in cluster  $i$, $ ped$ is the member locations or direction, and $C_t$ is the cluster locations or direction. The direction took a more crucial part since pedestrians can not be in a cluster if they move in a different direction.
  \subsubsection{Total Number of Pedestrians:} We compare the total number of pedestrians in the ground truth and the total number of pedestrians from all clusters for each frame.

We also evaluate the trajectory prediction using ADE and FDE metrics similar to other trajectory prediction methods \cite{Xugroupnet2022,Salzmann2020,Zhou2022}. ADE and FDE we used are the minimum ADE and FDE, means selecting the minimum score from the solution candidates.

\subsection{Implementation Details}
For the LOF method, we used a contamination parameter of 0.2 and a neighbourhood threshold of 80\% of the total cluster members. During the clustering process, based on our experiments, we applied thresholds of 120~px for distance and 50 degrees for direction for the best performance on our dataset.
ADE and FDE are evaluated with an 8-step history and 12-step  (short-term) predictions for all Algorithms. On SocialVAE, MART, DISTGCN, and random selection methods, we applied a 25-step history and 50-step (long-term) prediction to show clearly the accuracy difference; All methods were tested on difference machine and different configurations, which were adapted to each method. Therefore, our experiments are focused on how our methods behave on each method, rather than comparing the results on each method. Since no real-world coordinates are supplied by the crowd dataset, we use pixels as the unit of ADE and FDE.The execution time is in minutes, and the memory is in GiB for the unit. We do not include long-term prediction for Trajectron++ due to the high computational cost and poor results in early experiments. 

\subsection{Datasets and Track Extraction}

We use the MOT Head Tracking 21 dataset (MOT21) \cite{Sundararaman2021} for our trajectory dataset experiments.
Previous works on tracking have noted several problems with tracking: loss of tracking and identity (ID) switches, which can affect the downstream results on trajectory prediction \cite{Sundararaman2021,Aharon2022,Zhang2022}. TrajImput \cite{chib2024} declares these as missing values in observed trajectories and provides a trajectory dataset with imputed data. However, TrajImput also does not contain heavily crowded scenarios.

To measure the extent of these effects, we conducted a tracking and trajectory experiment with the Head Tracking 21 dataset using the HeadHunter-T \cite{Sundararaman2021} algorithm to show how much noise happens during tracking in crowded scenes. It is the only dataset with similar concerns on head tracking in dense crowds that works in multiple directions. We employed the crowd head dataset to train the detector, which is the crowd human dataset \cite{Shao2018} that is used by Headhunter (\cite{Sundararaman2021}) in the same way. This dataset consists of a mall dataset\cite{Loy2013} and a classroom dataset\cite{Peng2018}. Although Headhunter-T is not real-time, it can enhance accuracy from unclear images.

The results showed noticeable issues in performance, even for non-real-time object tracking (Table~\ref{tab:headhunter}). The highest ratios of correct detection, recall, and MOTA(Multiple Object Tracking Accuracy) were reached on the HT21-02 scene; meanwhile, the false negatives and ID switches were mostly in the HT21-04. The performance of this algorithm, therefore, does not show promise to act as a dataset for learning trajectory prediction; all metrics are far from perfect.

\begin{table}[t]
  \caption{Tracking evaluation for MOT HT21 Dataset. The table shows the results of tracking error. The least noise was obtained on the HT21-02 and the most noise was obtained on the HT21-04 scenario.}
  \label{tab:headhunter}
   \centering
  \begin{tabular}{cccccl}
  \hline 
   Scene & IDF1 \(\uparrow\) & Rcll \(\uparrow\) & MOTA \(\uparrow\) & FN \(\downarrow\) & IDs \(\downarrow\)\\
    \hline 
    02& \textbf{61.2}\%& \textbf{74.2}\%& \textbf{ 66.8\%} & 168313 & 1890\\
    03&43.9\% & 56.2\% & 51.3\% & 103562 & 2144\\
    04&49.14\% & 60.4\% & 53.67\% &\bfseries 61286 &\textbf{919}\\
    \hline 
\end{tabular}
\end{table} 

\subsection{Clustering Quality Results}

\begin{table}[t]
     \caption{Cluster evaluation based on automated tracking outputs (HT) and ground truth trajectories (GT). All HT scenarios exhibit consistently low CMDD values, indicating minimal average distance across all frames. The ncluster column shows the number of clusters created between one member and more than 2 members, and npeds shows the total number of pedestrians. }
  \label{tab:resev}
   \centering
  \begin{tabular}{c@{\hskip 3pt}c@{\hskip 4pt}c@{\hskip 4pt}c@{\hskip 4pt}c@{\hskip 4pt}c@{\hskip 5pt}c@{\hskip 4pt}l}
    \hline
      Data &  CMDD \(\downarrow\) & CTEO \(\downarrow\) & CTEL \(\downarrow\)  &  ncluster &  npeds \\ 
    \hline 
    HT-02 & 7.93 & 0.0& 0.0  & 308/460 & 1264\\
    GT-02 & 6.77 & 0.26& 0.01 & 267/378 & 602\\
    \hline
    HT-03 & 7.04 &0.23& 0.01& 443/882 & 2577\\ 
    GT-03 & 5.60 & 0.98& 0.02& 353/482 & 811 \\
    \hline
    HT-04 &7.99 & 0.31& 0.01& 281/653 & 1387\\
    GT-04 &8.084& 0& 0.0 & 270/367 & 580\\
  \hline
\end{tabular}
\end{table} 

\begin{table}[t]
  \caption{Trajectory prediction performance using Trajectron++ on HeadHunter.}
  \label{tab:trajectorypred}
   \centering
  \begin{tabular}{c@{\hskip 3pt}c@{\hskip 3pt}c@{\hskip 3pt}c@{\hskip 3pt}c@{\hskip 3pt}c@{\hskip 3pt}l@{\hskip 3pt}}
    \hline
   Scene&Source & Exec\(\downarrow\) & ADE \(\downarrow\)& FDE \(\downarrow\) & Train Data& nNode\\
   \hline
02&GT&2703.3&3.06&6.53&128,217&448\\
02&Cluster &\bfseries 672.4& 9.74&17.72& \bfseries62,396& 268\\
02&Tracking & 1496.5&\bfseries9.56&\bfseries17.40&102,852&805\\
    \hline
03&GT&1036.2&25.90&47.63& 161,198 & 616\\
03&Cluster&\bfseries 530.0&26.50&48.69 & \bfseries 60,539&528\\
03&Tracking&2572.0&\bfseries26.41&\bfseries48.66&110,729&1599\\
\hline
04&GT&1727.9&12.26&22.44&107,489&431\\
04&Cluster&\bfseries 342.6 &\bfseries12.28&\bfseries22.48&\bfseries51,053&427\\
04&Tracking&1197.6&12.32&22.58&77,830&853\\
\hline
\end{tabular}
\end{table}
\begin{table}[t]
  \caption{Trajectory prediction performance with SocialVAE. This table shows a significant execution time reduction and shows how the cluster outperforms the accuracy of the tracking results on 50 future steps and traditional random selection methods.}
  \label{tab:trajectorypredvae}
   \centering
  \begin{tabular}{l@{\hskip 3pt}l@{\hskip 3pt}l@{\hskip 3pt}l@{\hskip 3pt}l@{\hskip 3pt}l@{\hskip 3pt}l@{\hskip 3pt}l@{\hskip 3pt}}
    \hline
   Scene & Source & Exec\(\downarrow\) & A12 \(\downarrow\)& A50 \(\downarrow\) & F12 \(\downarrow\)& F50 \(\downarrow\) &  Mem\(\downarrow\)\\
\hline
02&GT&23.3& 1.83&9.90& 3.15&17.23& 70.26\\
02&Cluster&\bfseries 13.3 & 2.02 & \bfseries 9.51 &\bfseries3.35 & \bfseries16.89& \bfseries 23.64\\
02&Tracking&20.0&\bfseries 2.01 &12.18&3.43 &20.86&45.72\\
02&Random&13.3&-& 13.47&-&23.44&-\\
    \hline
03&GT&30.0 & 3.20 &15.97&4.802&34.08& 120.17\\
03&Cluster&\bfseries 13.3& 4.10&\bfseries 18.66 & 6.79 &\bfseries37.11& \bfseries 17.85\\
03&Tracking&26.7& \bfseries3.47 &21.89 & \bfseries 5.31 &40.49 & 25.18\\
03&Random& 13.3&-& 30.53&- & 58.75&-\\
\hline
04&GT&16.7 &  2.43&12.77& 4.36&24.82&45.04\\
04&Cluster&\bfseries 11.7  &2.95&\bfseries15.59& 5.62 &\bfseries35.39 &\bfseries15.99\\
04&Tracking&16.7&\bfseries2.72&20.19&\bfseries 5.032&38.26&28.01\\
04&Random& 11.7&-& 25.61 &-& 47.37&-\\
\hline
\end{tabular}
\end{table}
\begin{table}[t]
  \caption{Trajectory prediction performance with MART. This table shows a significant reduction in execution time and shows how the cluster outperforms the execution time and memory usage of the tracking results and traditional random selection methods.}
  \label{tab:trajectorypredMART}
   \centering
  \begin{tabular}{l@{\hskip 3pt}l@{\hskip 3pt}l@{\hskip 3pt}l@{\hskip 3pt}l@{\hskip 3pt}l@{\hskip 3pt}l@{\hskip 3pt}l@{\hskip 3pt}}
   \hline
   Scene&Source & Exec\(\downarrow\) & A12 \(\downarrow\)& A50 \(\downarrow\) & F12 \(\downarrow\)& F50 \(\downarrow\) &  Mem \(\downarrow\)\\
\hline
02&GT&748.4& 1.12&5.56& 1.95&8.61&33.92\\
02&Cluster&\bfseries 236.3 & 1.55 &6.22 &3.03&9.78 & \bfseries 15.79\\
02&Tracking&314.2&\bfseries1.51 &\bfseries5.99&\bfseries2.36&\bfseries9.32&25.30\\
02&Random& 136.5&-& 6.37&-&10.16&14.81\\
    \hline
03&GT&1485.2 & 1.89 &6.59&2.72&9.64& 54.66\\
03&Cluster&\bfseries 167.3 & 2.82& 9.77 &4.22&16.18 & \bfseries 20.69\\
03&Tracking&342.5&\bfseries2.29 &\bfseries8.89  & \bfseries3.18 &\bfseries15.00 & 26.23\\
03&Random&142.7&-& 10.66&- & 17.49&19.94\\
\hline
04&GT&377.9&  1.22&6.24& 2.20&9.12&22.91\\
04&Cluster&\bfseries 155.9&1.97&7.96 & 3.15 & 12.51  &\bfseries 8.28\\
04&Tracking&209.7&\bfseries1.75&\bfseries7.18&\bfseries2.84&\bfseries11.03&12.50\\
04&Random&127.2&-& 7.47 &-& 11.67&7.89\\
\hline
\end{tabular}
\end{table}

\begin{table}[t]
  \caption{Trajectory prediction performance with DSTIGCN. The table shows a significant reduction in execution time and Memory from the tracking result, and a slight reduction or increase in accuracy for both prediction horizons.}
  \label{tab:trajectorypredDSTIGCN}
  \centering
  \begin{tabular}{l@{\hskip 3pt}l@{\hskip 3pt}l@{\hskip 3pt}l@{\hskip 3pt}l@{\hskip 3pt}l@{\hskip 3pt}l@{\hskip 3pt}l@{\hskip 3pt}}
  \hline
  Scene & Source & Exec \(\downarrow\) & A12 \(\downarrow\) & A50 \(\downarrow\) & F12 \(\downarrow\) & F50 \(\downarrow\) & Mem\(\downarrow\)\\
  \hline
02 & GT       & 485.2 & 1.21 & 6     & 2.32 & 10.24  & 6.40\\
02 & Cluster  & \bfseries 126.2 & 1.65 & 6.632 & 3.50 & 11.288 & \bfseries 2.98\\
02 & Tracking & 183.4 & \bfseries 1.62 & \bfseries 6.42 & \bfseries 2.77 & \bfseries 10.952 & 4.78\\
02 & Random   & 84.0  & - & 10.648 & - & 18.592 & 2.80\\
\hline
03 & GT       & 961.5 & 2.75 & 9.6   & 4.78 & 16.936 & 10.32\\
03 & Cluster  & \bfseries 161.3 &\bfseries 3.10 & \bfseries 11.584 &\bfseries  4.74 & \bfseries21.512 & \bfseries 3.91\\
03 & Tracking & 231.7 &  3.41 &  12.04 &  5.05 & 22.384 & 4.95\\
03 & Random   & 185.6 & - & 8.528 & - & 13.992 & 3.76\\
\hline
04 & GT       & 269.7 & 1.34 & 6.856 & 2.78 & 11.536 & 4.32\\
04 & Cluster  & \bfseries 60.6 & 1.92 & 8.152 & 3.61 & 14.752 & \bfseries 1.56\\
04 & Tracking & 122.1 & \bfseries 1.94 & \bfseries 7.968 & \bfseries 3.77 & \bfseries 14.64 & 2.36\\
04 & Random   & 58.4 & - & 11.344 & - & 20.8208 & 1.49\\
\hline
\end{tabular}
\end{table}
Our first set of experiments evaluates how well our method is able to cluster ground-truth and tracker-generated pedestrian trajectories. We see that GT data perform similarly to HT data, even though HT data has tracking noise (Table~\ref{tab:resev}). $CMDD$ shows the membership score on the clusters that have more than one member(eq.~\ref{eq:CMDD}). The CMDD shows that our method could maintain the membership score as similar as possible. The small average number shown in Table~\ref{tab:resev} for $CMDD1$ and $\mu CMDD2$ suggests that the number of pedestrian directions is not sensitive to outliers; it is accurately represented by the cluster, outweighing the number of outlier deviations. The very low CTEL (eq.~\ref{eq:CTEL}) and CTEO (eq.~\ref{eq:CTEO}) values indicate that the cluster trajectory is smooth without any anomalous displacement of the trajectory from one location to another. 

Npeds on GT is always lower than its HT since it does not suffer from tracking loss and re-identification processes. This also affects the creation of the cluster, whereas HT has more clusters than its ground truth. The initial nested cluster could potentially generate clusters with a single member if the pedestrian moves in the same direction but is situated at a far distance. 

\subsection{Trajectory Prediction Results}
\label{subsec:exp-results-traj-pred}
 
Our second set of experiments measures how clustering affects trajectory prediction, in terms of execution time, accuracy (ADE and FDE), and memory usage.
We used 3 scenarios with 3 different data sources for each: ground truth (GT), Cluster Centroids, and tracking from HeadHunter. To evaluate the cluster for truth trajectory data fairly, all the data sources shared the same evaluation data from the ground truth. 
Results with Trajectron++, SocialVAE, MART, and DSTIGCN are given in Tables \ref{tab:trajectorypred}, \ref{tab:trajectorypredvae}  \ref{tab:trajectorypredMART}, and \ref{tab:trajectorypredDSTIGCN}. We fed the trajectory predictors input with the centroids output by our clustering; therefore, the cluster substitutes the pedestrians. Our proposed clustering approach significantly reduced execution time (from 33.33\% to 79.4\%) and reduced its maximum memory usage up to 42.93\%, while maintaining robust accuracy in trajectory tracking for all trajectory prediction algorithms compared with the tracking data source, with only a slight decline in ADE and FDE for 12-step predictions, indicating effective data representation from the tracking process. Notably, in 50-step predictions, SocialVAE with clustering outperforms its tracking-based model and slightly surpasses ground-truth-based accuracy, as shown in Table \ref{tab:trajectorypredvae}. In MART and Trajectron++, our clustering slightly reduces accuracy but significantly improves maximum memory usage and execution time. 

For the long-term results in Tables \ref{tab:trajectorypredvae}, \ref{tab:trajectorypredMART}, and \ref{tab:trajectorypredDSTIGCN} we also include a baseline approach that subsamples the raw pedestrian data, deciding at random whether to include each individual's full trajectory; this provides a naive alternative to our clustering that can also speed up trajectory prediction.
In all cases, our clustering approach surpasses the accuracy of this traditional random selection.
Overall, our clustering approach demonstrates a varied, slight reduction or improvement in accuracy and consistently outperforms random sampling while drastically reducing execution time and memory usage compared to using the raw pedestrian data. 

\section{Conclusion}
In this paper, we presented our novel dynamic clustering approach for faster and more efficient pedestrian trajectory prediction in dense-crowd situations. It significantly reduces execution time and memory usage while maintaining accuracy with minimal performance degradation on short-term prediction and even performance improvement on long-term prediction compared to the tracking input and random selection. This work also demonstrates how our proposed clustering strategy accurately represents pedestrians based on their dynamic membership distance while maintaining the smoothness and continuity of the trajectory. This research lays the groundwork for future investigations into real-time prediction by simplifying dense crowd trajectory data.

\printcredits

\bibliographystyle{cas-model2-names}

\bibliography{cas-refs}


\end{document}